\pdfoutput=1
\documentclass[conference]{IEEEtran}
\usepackage[utf8]{inputenc}
\usepackage[T1]{fontenc}
\usepackage{amsmath,amssymb}
\usepackage{graphicx}
\usepackage{booktabs}
\usepackage{cite}
\graphicspath{{figures/}}

\begin{document}

\title{A Belief-State World Model for Catheter Navigation under Sparse Fluoroscopy: A Planar Proof of Concept}

\author{\IEEEauthorblockN{Damini Rijhwani}
\IEEEauthorblockA{Automation Core Inc.\\
Cambridge, MA, USA\\
damini@automationcoreinc.com\\
ORCID: 0009-0000-2253-6111}}

\maketitle

\begin{abstract}
Endovascular catheter navigation relies on continuous fluoroscopy, exposing patients and clinical staff to ionizing radiation throughout the procedure. We investigate whether a physics-based world model can sustain the navigation task between deliberately sparse X-ray acquisitions, and whether the model can signal when its internal state estimate is no longer reliable. We formulate sparse-fluoroscopy navigation as a partially observable Markov decision process where a Cosserat rod simulator supplies transition dynamics, sparse noisy projections provide observations, and a particle filter maintains a belief over the device state, reduced in this implementation to a tip state with a geometric contact proxy. We evaluate this formulation in a deliberately simplified setting: a synthetic planar vessel phantom with a quasi-static rod model and simulated projections, without clinical or animal data. In this setting, the belief-state model tracks the simulated tip with a root mean square error of 1.30 mm while acquiring observations at one fifteenth of the continuous rate (0.97 mm at every frame, 1.51 mm at one thirtieth), reported 90 percent credible intervals achieve 0.92 empirical coverage, and the belief-derived contact risk estimate discriminates unsafe contact events with an AUROC of 0.68. These results constitute a proof of concept on a simplified simulation rather than a demonstration of clinical readiness; their purpose is to establish that calibrated belief, rather than point-estimate accuracy alone, is the essential property a sparse-imaging world model must deliver.
\end{abstract}


\section{Introduction}

Endovascular interventions are guided almost entirely by fluoroscopy. Clinicians advance a guidewire or catheter while viewing a continuous X-ray stream, running the imaging system throughout much of the procedure. The resulting radiation burden presents a recognized occupational hazard for interventional teams and a dose concern for patients; professional guidelines advise operators to keep exposure as low as reasonably achievable and treat every second of beam-on time as a cost \cite{miller2010cirse}. Current dose-reduction practices center on pulsed acquisition, collimation, and operator discipline. The imaging schedule itself, however, remains essentially continuous because operators lack alternative sources of information regarding device location.

Robotics provides a candidate for that missing information source. A world model is an internal simulation of the environment that an agent rolls forward to predict action consequences, querying real observations only to correct internal drift \cite{ha2018worldmodels, hafner2023nature}. The components for a physical catheter world model already exist across separate literatures. Cosserat rod theory provides a mechanically grounded model of slender flexible instruments under tendon actuation and external loading \cite{rucker2011tro}, and has been applied to intraoperative contact-force estimation \cite{back2015iros}. Open simulation frameworks support interactive rod and contact mechanics for medical scenarios \cite{allard2007sofa}. Deep networks segment catheters in fluoroscopy frames in real time \cite{ambrosini2017tmi}, enabling acquired frames to serve as direct measurements. Cardiovascular digital-twin research emphasizes exactly this form of patient-specific predictive model \cite{coorey2022digitaltwin}. In principle, procedures could rely primarily on internal simulation, querying X-rays sparsely at low frame rates or on demand to reset accumulated error.

What is missing is not forward prediction, but calibrated belief. Between observations, the internal model operates open-loop, with error accumulating from contact, friction, and actuation noise. A system providing an operator with a confident but inaccurate tip position, or an incorrect wall contact force, is worse than no assistance. The fundamental question for a sparse-imaging world model is therefore whether it recognizes when its estimate becomes unreliable: whether its reported uncertainty is statistically honest, and whether safety metrics derived from the belief, such as contact-force threshold exceedance, remain informative as observation frequency decreases.

This paper takes an initial, focused step toward addressing that question. Our contributions are:
\begin{itemize}
  \item a POMDP formulation of catheter navigation under sparse fluoroscopy, where the transition model is a Cosserat rod simulator, the observation model is a sparse noisy projection, and safety is evaluated as a contact-force exceedance probability under the belief;
  \item a planar proof-of-concept implementation pairing a quasi-static rod simulator with a particle-filter belief over device state, reduced here to a tip state with a geometric contact proxy;
  \item a calibration-focused evaluation protocol reporting tip tracking error (1.30 mm at one fifteenth of continuous frame rate), empirical coverage of reported credible intervals (0.92), and unsafe contact event discrimination (AUROC 0.68) across reduced observation rates.
\end{itemize}

We state the experimental scope directly. All evaluations run on a synthetic planar vessel phantom with a simplified quasi-static rod model; there are no clinical or animal data, no real fluoroscopic streams, and no human operator in the loop. These results do not constitute a claim of clinical readiness. Our goal is to establish the formulation and evaluation protocol at a focused scale where calibration can be assessed rigorously, providing a clear foundation for future work on complex anatomy and imaging.

\section{Method}

\subsection{Problem Formulation}

We model navigation under sparse fluoroscopy as a partially observable Markov decision process specified by the tuple $W = (S, A, T, O, Z, R)$. Each element is instantiated concretely for the planar simulator used in this study.

\textbf{State $S$.} A state $\mathbf{s}_t$ collects the discretized rod configuration, the node positions and tangent angles of an $N$-node planar rod, alongside per-node contact indicators and penetration depths at active contacts. Vessel phantom geometry is fixed and known, entering the model as a parameter rather than part of the state.

\textbf{Action $A$.} An action $\mathbf{a}_t = (\Delta \ell_t, \Delta \phi_t)$ represents the proximal control input: insertion increment $\Delta \ell_t$ along the sheath axis and base angle increment $\Delta \phi_t$. This is the planar reduction of insertion, rotation, and tendon inputs on steerable devices \cite{rucker2011tro}.

\textbf{Transition $T$.} Given $(\mathbf{s}_t, \mathbf{a}_t)$, the next state is determined by applying the perturbed base action and solving quasi-static energy minimization for rod equilibrium, incorporating elastic bending energy, vessel wall penalty contact energy, and additive Gaussian actuation noise. The transition is stochastic through actuation noise, while underlying mechanics remain deterministic.

\textbf{Observation $O$ and operator $Z$.} When acquired, an observation is a sparse noisy projection of the rod: positions of a subset of nodes projected onto a one-dimensional detector axis with Gaussian pixel noise, representing segmented catheter centerlines in fluoroscopy \cite{ambrosini2017tmi}. Operator $Z$ yields likelihood $p(\mathbf{o}_t \mid \mathbf{s}_t)$ under this projection and noise model. Sparsity operates on two axes: frames arrive every $k$ steps, and each frame captures only a portion of the rod.

\textbf{Risk $R$.} Rather than a shaped reward, we track a safety functional: the probability, under the current belief, that the largest contact-force magnitude along the rod exceeds a fixed threshold $F_{\max}$,
\begin{equation}
R(b_t) \;=\; \Pr_{\mathbf{s} \sim b_t}\!\left[\, \max_{i} \lVert \mathbf{F}_{c,i}(\mathbf{s}) \rVert \;>\; F_{\max} \right],
\label{eq:risk}
\end{equation}
where $b_t$ is the state belief and $\mathbf{F}_{c,i}$ is the penalty contact force at node $i$. This exceedance probability, rather than a point estimate of force, is evaluated for discriminating unsafe events.

\subsection{Rod Mechanics}

The general formulation behind the transition model is Cosserat rod theory, which describes a slender instrument by a centerline $\mathbf{r}(s,t)$ and cross-section orientation $\mathbf{R}(s,t)$ over arc length $s$, balancing linear and angular momentum under internal loads and distributed external forces \cite{rucker2011tro}:
\begin{equation}
\begin{aligned}
\partial_s \mathbf{n} + \mathbf{f} &= \rho A \, \partial_{tt} \mathbf{r}, \\
\partial_s \mathbf{m} + \partial_s \mathbf{r} \times \mathbf{n} + \mathbf{l} &= \rho \mathbf{I} \, \partial_t \boldsymbol{\omega} + \boldsymbol{\omega} \times \rho \mathbf{I} \boldsymbol{\omega},
\end{aligned}
\label{eq:cosserat}
\end{equation}
where $\boldsymbol{\omega}(s,t)$ is the angular velocity of cross-section frame $\mathbf{R}(s,t)$, with internal force $\mathbf{n}$ and moment $\mathbf{m}$ linked to shear-extension and bending-torsion strains via linear constitutive relations with stiffness matrices $\mathbf{K}_{se}$ and $\mathbf{K}_{bt}$. Vessel interaction is represented by penalty contact forces at penetrating nodes with Coulomb friction on tangential components, standard in interactive medical simulation \cite{allard2007sofa, back2015iros}. The implementation replaces Coulomb friction with a quadratic displacement tether at previously loaded contacts, weighted by prior normal force, resisting sliding in a stick-like manner without saturating at Coulomb limit $\mu \lVert \mathbf{F}_{n} \rVert$.

The planar implementation incorporates four specific simplifications of \eqref{eq:cosserat}. First, it is quasi-static: inertial terms are omitted and each transition solves for static equilibrium, reflecting slow guidewire advancement while removing dynamic effects. Second, it is planar: the rod is inextensible and shear-free with pure bending, omitting torsion and out-of-plane motion. Third, contact relies on fixed-stiffness penalty forces without stick-slip friction, approximating tangential resistance. Fourth, insertion is handled in shape space rather than the material frame: new segments appear at the tip with the preceding tip angle, and friction references attach to shape nodes rather than material points, leaving axial material sliding unresolved. These choices preserve simulator speed for particle rollouts and keep modeling assumptions explicit, while setting aside physical details required for clinical systems.

\subsection{Belief Maintenance}

The belief $b_t$ is represented by $M$ weighted particles. Between observations, particles propagate through a stochastic transition model: $T$ in the general formulation and, in this planar implementation, a reduced kinematic surrogate of $T$ over tip state (position, heading, inserted length; Sec.~III). The particle cloud disperses at a rate governed by actuation and model-mismatch noise and by contact events, which introduce strong nonlinearities that challenge Gaussian filters. When an observation arrives, particle weights update via likelihood $Z$, with inflated noise scales and centerline-based marker predictions absorbing surrogate shape errors, followed by resampling. This reflects the standard prediction-correction assimilation pattern of world-model agents \cite{ha2018worldmodels, hafner2023nature}, implemented here with mechanical rather than learned transitions. All reported metrics, tip estimates, credible intervals, and risk functionals in \eqref{eq:risk}, derive from this particle set, evaluating belief calibration directly.

\section{Experiments and Results}

\subsection{Setup}

\textbf{Phantom and rod.} The environment is a synthetic planar vessel: a cubic-spline centerline with two bends and arc-varying radius, including a Gaussian narrowing with minimum lumen radius of 1.6 mm, represented as a signed distance field. The instrument is a discrete elastic rod with $N = 40$ segments of length 3 mm and radius 0.6 mm; each control step solves quasi-static equilibrium via L-BFGS energy minimization with analytic gradients (bending, base clamp, penalty wall contact, and tangential resistance).

\textbf{Episodes and observations.} Each episode comprises a scripted 300-step insertion with nominal 0.35 mm advance per step plus actuation perturbations. The simulated C-arm projects the tip and 8 radio-opaque markers onto a 1D detector at 35 degrees with 0.2 mm per pixel and Gaussian noise of 1.5 pixels (0.3 mm). The tracker receives every $k$-th frame for $k \in \{1, 5, 15, 30\}$; assuming a 15 Hz baseline, these correspond to 15 Hz down to one frame every two seconds.

\textbf{Trackers.} The particle filter (PF) is a bootstrap filter with $M = 500$ particles. In this planar implementation, each particle tracks tip position, heading, and inserted length, representing a reduced parameterization sufficient for the evaluated metrics. The transition applies commanded actions with inflated process noise and a weak centerline guidance term, enforces a hard lumen constraint, and uses systematic resampling with roughening. The dead-reckoning (DR) baseline propagates the identical transition model without assimilation, isolating the contribution of sparse observations. The contact-risk predictor uses a location proxy for \eqref{eq:risk}: the fraction of particles contacting the wall, evaluated against ground-truth events defined as maximum nodal contact force exceeding the 90th percentile of the run.

\textbf{Protocol.} We evaluate 20 fixed seeds per $(k, \text{tracker})$ condition, reporting tip RMSE (mean $\pm$ std across seeds), empirical coverage of the 90 percent credible region (summarized by weighted particle mean and covariance, using the Gaussian ellipse at the chi-square 90 percent threshold for two degrees of freedom, measuring the fraction of steps where the true tip falls within this region), and contact predictor AUROC pooled across seeds. The sweep completes in 55 s on a laptop CPU with numpy/scipy, reproducing bit-identically.

\begin{figure}[t]
\centering
\includegraphics[width=\columnwidth]{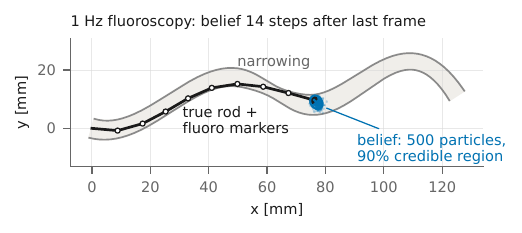}
\caption{Simulation snapshot on the synthetic planar phantom (seed 0, $k=15$, step 194, which is 14 steps after the last acquired frame). Ground-truth rod, particle cloud, and dead-reckoning estimate inside the synthetic vessel. This is a proof-of-concept simulation; no real anatomy or fluoroscopy is involved.}
\label{fig:snapshot}
\end{figure}

\begin{figure}[t]
\centering
\includegraphics[width=\columnwidth]{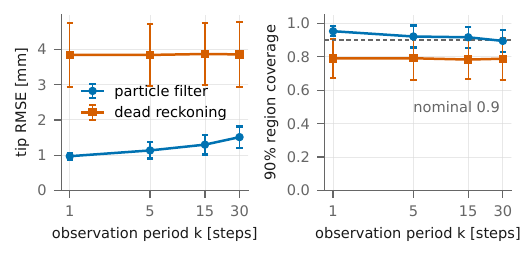}
\caption{Tip RMSE, 90 percent coverage, and contact AUROC versus observation interval $k$ on the synthetic planar phantom (20 seeds). Proof-of-concept results; error bars are standard deviations across seeds.}
\label{fig:metrics}
\end{figure}

\begin{table}[t]
\caption{Results on the synthetic planar phantom (20 seeds). Tip RMSE in mm (mean $\pm$ std across seeds); Cov90 = empirical coverage of the 90\% credible region, mean $\pm$ std across seeds (nominal 0.90); AUROC pooled across seeds.}
\label{tab:results}
\centering
\begin{tabular}{clccc}
\toprule
$k$ & Tracker & Tip RMSE (mm) & Cov90 & AUROC \\
\midrule
1  & PF & $0.97 \pm 0.09$ & $0.953 \pm 0.029$ & 0.700 \\
   & DR & $3.84 \pm 0.91$ & $0.790 \pm 0.119$ & 0.707 \\
\midrule
5  & PF & $1.13 \pm 0.23$ & $0.921 \pm 0.066$ & 0.721 \\
   & DR & $3.84 \pm 0.87$ & $0.791 \pm 0.130$ & 0.700 \\
\midrule
15 & PF & $1.30 \pm 0.28$ & $0.917 \pm 0.062$ & 0.684 \\
   & DR & $3.87 \pm 0.89$ & $0.783 \pm 0.115$ & 0.702 \\
\midrule
30 & PF & $1.51 \pm 0.30$ & $0.895 \pm 0.065$ & 0.668 \\
   & DR & $3.86 \pm 0.92$ & $0.788 \pm 0.125$ & 0.703 \\
\bottomrule
\end{tabular}
\end{table}

\subsection{Tracking under Sparsification}

Table~\ref{tab:results} and Fig.~\ref{fig:metrics} detail the sweep results. The particle filter degrades smoothly as observation frequency falls: tip RMSE rises from $0.97 \pm 0.09$ mm at $k=1$ to $1.51 \pm 0.30$ mm at $k=30$. Dead reckoning remains near 3.84 to 3.87 mm across all $k$; without assimilation, its error reflects actuation noise and lumen geometry, with the lumen constraint bounding maximum drift. The belief-state tracker maintains errors roughly 2.5 to 4 times lower than the open-loop baseline across all observation rates. Fig.~\ref{fig:snapshot} shows a representative state 14 steps after the last frame: the particle cloud elongates along the direction of motion, capturing appropriate uncertainty growth between observations.

\subsection{Calibration}

Coverage represents the central finding. The PF 90 percent credible region covers the true tip in 0.953 of steps at $k=1$ and 0.895 at $k=30$ (Table~\ref{tab:results}), close to the nominal 0.90 target at every sparsity level; the filter widens its reported uncertainty at roughly the rate its actual error grows. Dead reckoning, with identical process noise, under-covers at 0.78 to 0.79 across all $k$, underestimating true error. This distinction is fundamental: both models output estimates and uncertainty, but only the assimilating belief maintains calibration, a requirement for operators deciding whether additional imaging is needed.

\subsection{Contact-Event Discrimination}

The location-based contact proxy achieves pooled AUROC values between 0.668 and 0.721 for the PF across $k$ (Table~\ref{tab:results}), compared to 0.700 to 0.707 for dead reckoning. Two observations follow. First, the predictor is informative above chance at every sparsity level, so even a crude belief-derived contact signal carries some safety-relevant information. Second, it does not separate PF from dead reckoning: because the proxy relies only on position relative to the wall and both trackers stay within the lumen, improved localization does not improve event discrimination in this planar setting. Evaluating the full risk functional of \eqref{eq:risk} requires particles tracking mechanical state (contact force rather than position), which this reduced parameterization omits. We report this negative result directly, identifying force-aware belief states as the key requirement for future development.

\section{Discussion and Limitations}

The experiments support one narrow claim: in controlled planar simulation, a particle-filter belief over a mechanical catheter model remains near-calibrated as fluoroscopy is sparsified thirty-fold, whereas an open-loop model becomes overconfident. Calibration, rather than RMSE alone, separates the two approaches, providing the mechanism for a sparse-imaging system to request frames when uncertainty grows.

The limitations are substantial and acknowledged directly. The phantom is synthetic and planar; real vasculature is three-dimensional, patient-specific, and deformable, whereas our wall model is rigid. The rod model is quasi-static with pure bending, omitting dynamic effects, torsion, out-of-plane deformation, and stick-slip friction. The projection model uses a 1D detector with Gaussian noise, omitting fluoroscopic physics, segmentation errors, and occlusion. The particle state tracks tip parameters rather than the full rod, preventing the contact proxy from benefiting from assimilation. The study includes no clinical or animal data and no human operator; the scripted policy does not adapt to belief states. These values do not predict clinical performance. What transfers is the formulation and evaluation protocol: future systems can be assessed using the same metrics of error, coverage, and event discrimination.

\section{Conclusion}

We formulated catheter navigation under sparse fluoroscopy as a POMDP with a Cosserat mechanical transition model, showing in a planar synthetic proof of concept that particle-filter beliefs remain near-calibrated (coverage 0.89 to 0.95 vs nominal 0.90) while tracking tip position at 0.97 to 1.51 mm across a thirty-fold reduction in observation rate. A related direction is aggregating this belief-derived risk functional across simulated patient populations, so that exceedance probabilities support cohort-level safety analysis rather than single-procedure monitoring, connecting belief states to active planning \cite{greigarn2019tbme}. Extending this twin to volumetric anatomy with 3D Cosserat rods, deformable vessel walls, force-aware particle states, and realistic imaging models represents the essential next step before claims beyond proof of concept can be made.

\bibliographystyle{IEEEtran}
\bibliography{refs_verified}

\end{document}